\documentclass{isprs} 
\usepackage{subfigure}
\usepackage{setspace}
\usepackage{geometry} 
\usepackage{epstopdf}
\usepackage[labelsep=period]{caption}  
\usepackage[british]{babel} 
\usepackage[hang]{footmisc}

\usepackage{multirow}
\usepackage{xcolor}
\usepackage{booktabs}
\usepackage{hyperref}
\usepackage{cite}

\usepackage[authoryear]{natbib}

\newcommand{\rfname}{REFERENCES}
\makeatletter
\renewcommand{\bibsection}{%
   \section*{\rfname%
            \@mkboth{\MakeUppercase{\rfname}}{\MakeUppercase{\rfname}}%
   }
}
\makeatother

\usepackage{array}
\newcolumntype{L}{>{\centering\arraybackslash}m{0.125\linewidth}}

\begin{document}

\title{DO WE STILL NEED IMAGENET PRE-TRAINING IN REMOTE SENSING SCENE CLASSIFICATION?}
\date{}

\author{Vladimir Risojevi\'{c}\thanks{Corresponding author}, Vladan Stojni\'{c}}

\address{Faculty of Electrical Engineering\\University of Banja Luka\\Banja Luka, Bosnia and Herzegovina\\\{vladimir.risojevic, vladan.stojnic\}@etf.unibl.org}


\icwg{}   

\abstract{
Due to the scarcity of labeled data, using supervised models pre-trained on ImageNet is a de facto standard in remote sensing scene classification. Recently, the availability of larger high resolution remote sensing (HRRS) image datasets and progress in self-supervised learning have brought up the questions of whether supervised ImageNet pre-training is still necessary for remote sensing scene classification and would supervised pre-training on HRRS image datasets or self-supervised pre-training on ImageNet achieve better results on target remote sensing scene classification tasks. To answer these questions, in this paper we both train models from scratch and fine-tune supervised and self-supervised ImageNet models on several HRRS image datasets. We also evaluate the transferability of learned representations to HRRS scene classification tasks and show that self-supervised pre-training outperforms the supervised one, while the performance of HRRS pre-training is similar to self-supervised pre-training or slightly lower. Finally, we propose using an ImageNet pre-trained model combined with a second round of pre-training using in-domain HRRS images, i.e. domain-adaptive pre-training.  The experimental results show that domain-adaptive pre-training results in models that achieve state-of-the-art results on HRRS scene classification benchmarks. The source code and pre-trained models are available at \url{https://github.com/risojevicv/RSSC-transfer}.}


\maketitle


\section{INTRODUCTION}

Transfer learning opened up the possibility of applying deep learning to domains in which labeled data is scarce, difficult or expensive to acquire, such as remote sensing \citep{ball2018state}. A standard approach for applying transfer learning in high resolution remote sensing (HRRS) scene classification has been to start with a supervised model trained on ImageNet and either use it for feature extraction or fine-tune it to the target task \citep{hu2015transferring, nogueira2017towards}. We refer to the first round of training as pre-training. In the case of feature extraction the pre-trained model is used to compute the features of the target images, which can subsequently be used e.g., for training a classifier or for image retrieval. On the other hand, in fine-tuning the entire network is optimized for the target classification task.

The most prominent benchmark in HRRS scene classification for a long time has been UCM dataset with 21 classes and only 100 images per class \citep{yang2010bag}, which is not enough for training a model from scratch. However, in the previous years, several larger datasets of HRRS images have appeared with a goal of establishing new benchmarks for HRRS scene classification \citep{xia2017aid, cheng2017remote, zhou2018patternnet, li2020rsi, qi2020mlrsnet}. These datasets contain more classes and more images per class than UCM and, having in mind the domain difference between everyday objects in ImageNet and remote sensing scenes, a question arises whether we still benefit from ImageNet pre-training in HRRS scene classification. 

On the other hand, ImageNet has been successfully used for self-supervised pre-training and it was recently shown in \citep{ericsson2021well} that self-supervised models transfer better than supervised models to a number of downstream tasks. Therefore, to obtain a complete picture of usefulness of ImageNet pre-training in remote sensing scene classification self-supervised pre-training should also be included in the analysis.

To answer the titular question we both trained from scratch and fine-tuned ImageNet pre-trained supervised and self-supervised convolutional neural networks on multiple HRRS image datasets and compared the resulting classification accuracies. Furthermore, we examined whether representations learned on HRRS image datasets transfer better to other HRRS scene classification tasks compared to the representations learned on ImageNet. We used multiple source and target HRRS datasets of various sizes and class distributions with a goal of finding the factors that influence the model transferability.

In addition, it has recently been observed that transfer performance of large unsupervised pre-trained language models can be improved by additional unsupervised pre-training on data from the target domain, an approach known as domain-adaptive pre-training \citep{gururangan2020don}. Similar results with self-supervised and supervised learning for several computer vision tasks were presented in \citep{reed2021self}. Motivated by these results, we also investigated if domain-adaptive pre-training can improve upon pre-training on only ImageNet or HRRS images. In contrast to \citep{gururangan2020don}, we experiment with supervised domain-adaptive pre-training and show that it can improve the classification accuracies on the remote sensing test datasets without any additional model complexity. The resulting performances are similar or better than state-of-the-art obtained using architectural modifications. Therefore, our results can be regarded as the new baselines and models reused as backbones in all applications where ImageNet pre-trained supervised networks have been used.

The main contributions of this paper are:
\begin{enumerate}
\item Comparison of training from scratch and fine-tuning ImageNet pre-trained supervised and self-supervised convolutional neural networks on HRRS image datasets.
\item Evaluation of transferability of representations learned from scratch on HRRS image datasets to other HRRS scene classification tasks.
\item Introduction of supervised domain-adaptive pre-training to HRRS scene classification and empirical analysis of the factors contributing to its good performance.
\item Publishing the source code and pre-trained models to facilitate new research and applications of transfer learning.
\end{enumerate}

The rest of the paper is organized as follows. In Section~\ref{sec:related} related work is reviewed. The datasets and training details are described in Section~\ref{sec:mm}. The experimental results are presented in Section~\ref{sec:experiments}. We discuss the results in Section~\ref{sec:discussion} and Section~\ref{sec:conclusion} concludes the paper.

\section{RELATED WORK}
\label{sec:related}

It has long been known that networks pre-trained on ImageNet can produce representations suitable for different target tasks \citep{sharif2014cnn, donahue2014decaf}. Consequently, the question of factors influencing the transferability of the learned features had been investigated in \citep{yosinski2014transferable, azizpour2015factors, huh2016makes}. It was shown that  similarity of the source and target tasks, as well as diversity and size of the source dataset influence the performance on a target task. However, these results are not consistent across the source and target tasks, and recently in \citep{he2019rethinking} was observed that ImageNet pre-training did not improve object detection and instance segmentation performance. Additionaly, it was found in \citep{kornblith2019better} that the networks pre-trained on ImageNet did not improve performance on target tasks from significantly different domains, requiring fine-grained classification or having more training data. These results spurred the interest in examining the effects of ImageNet pre-training in medical imaging \citep{raghu2019transfusion, ke2021chextransfer, hosseinzadeh2021systematic}, which shares some of the problems, such as the lack of large labeled datasets and need for domain experts for labeling, with remote sensing. However, to the best of our knowledge, a systematic investigation of ImageNet pre-training in HRRS scene classification is still missing.

Models pre-trained on ImageNet have quickly gained popularity for remote sensing image classification \citep{penatti2015deep, hu2015transferring, marmanis2015deep, liang2016transfer, nogueira2017towards, zhao2017transfer, tong2020land}. Although pre-training on ImageNet is a de facto standard, several papers also reported experiments with pre-training on remote sensing image datasets. In \citep{chen2019improved} a HRRS scene classification model pre-trained on a dataset constructed taking a union of RESISC45, PatternNet, and RSI-CB showed slightly improved results on UCM classification compared to ImageNet pre-training. In contrast, an analysis in \citep{pires2020convolutional} showed that the networks pre-trained on ImageNet outperform the ones pre-trained on PatternNet in transfer learning to AID and UCM datasets. Concurrently with this work, Million-AID, the largest HRRS image dataset to date, has been published along with the experiments showing that pre-training on Million-AID results in better transfer learning performance on RESISC45 and AID than pre-training on ImageNet \citep{long2022aerial}.

The work in \citep{neumann2020training} is similar to ours in the sense that transfer from both ImageNet and remote sensing image datasets is explored. The experiments had been performed on three medium-resolution (BigEarthNet \citep{sumbul2019bigearthnet}, EuroSAT \citep{helber2019eurosat}, and So2Sat \citep{zhu2019so2sat}) as well as two high-resolution datasets (RESISC45 and UCM) and the results showed that fine-tuning the models pre-trained on remote sensing datasets resulted in better classification accuracies than fine-tuning the models pre-trained on ImageNet. Furthemore, multiresolution datasets led to more transferable representations and medium-resolution datasets did not yield good generalization to high-resolution datasets. However, since only two HRRS image datasets were used, the question of the factors that influence transferability for HRRS scene classification remained unanswered. In this paper we perform the experiments on multiple HRRS image datasets of various sizes and with different numbers of classes with a goal of identifying the factors that influence the transferability of the obtained representations the most. 

Self-supervised learning holds a promise to reduce the need for large labeled datasets in training deep learning models \citep{jing2020self}. Recently, it has been shown that the best self-supervised ImageNet models can outperform supervised ImageNet models in transferring to various downstream tasks \citep{ericsson2021well}. Although in most cases ImageNet or larger datasets of images of everyday objects and scenes are used for self-supervised pre-training, there are also attempts to use remote sensing images for that purpose \citep{manas2021seasonal, ayush2021geography, stojnic2021self}. However, since the preliminary experiments with pre-trained models from \citep{manas2021seasonal} resulted in poor performances, we did not use remote sensing images for self-supervised pre-training.

\section{MATERIALS AND METHODS}
\label{sec:mm}

We used six HRRS image datasets: MLRSNet \citep{qi2020mlrsnet}, RESISC45 \citep{cheng2017remote}, PatternNet \citep{zhou2018patternnet}, RSI-CB \citep{li2020rsi}, AID \citep{xia2017aid}, and UCM \citep{yang2010bag}. The main properties of these datasets are listed in Table~\ref{tab:datasets}. The first five datasets were used as both source and target datasets while UCM was used only as the target dataset because it is too small for training a network from scratch. MLRSNet can be used for training both single-label and multi-label classifiers so we included both scenarios in our experiments. We also used ImageNet-1k and ImageNet-100 as source datasets. ImageNet-100 is a subset of ImageNet-1k with 100 classes and total of 131,689 images \citep{tian2020contrastive}. Having the size similar to MLRSNet, ImageNet-100 enabled us to assess how the source domain influences representation transferability. With ImageNet-1k as the source dataset we experimented with both supervised and self-supervised pre-training. For self-supervised pre-training we chose SwAV \citep{caron2020unsupervised}, which had shown a good transfer performance in \citep{ericsson2021well}. We did not train networks on ImageNet-1k ourselves and rather used the pretrained supervised model available in Keras as well as the PyTorch implementation and weights of a self-supervised model pre-trained using SwAV and provided by the authors of \citep{caron2020unsupervised}. 

 \begin{table*}
     \centering
     \begin{tabular}{lrrrrr}
          \toprule
          Dataset & Size & Classes & Image size & Resolution (m) & Annotations \\
          \midrule
          MLRSNet \citep{qi2020mlrsnet}          & 109,161 & 46/60 & $256 \times 256$ & 0.1 - 10 & single/multi-label \\
          RESISC45 \citep{cheng2017remote}  &  31,500 &  45   & $256 \times 256$ & 0.2 - 30  & single-label \\
          PatternNet \citep{zhou2018patternnet}  &  30,400 &  38   & $256 \times 256$ & 0.062 - 4.693  & single-label \\
          RSI-CB \citep{li2020rsi}               &  24,000 &  35   & $256 \times 256$ & 0.22 - 3  & single-label \\
          AID \citep{xia2017aid}                 &  10,000 &  30   & $600 \times 600$ & 0.5 - 8  & single-label \\
          UCM \citep{yang2010bag}                &   2,100 &  21   & $256 \times 256$ & 0.3  & single-label \\
          \bottomrule
     \end{tabular}
     \caption{Details of HRRS image datasets used in the experiments.}
     \label{tab:datasets}
 \end{table*}

For each source dataset, we used 80\% of images from each class for training/fine-tuning and the rest for testing. We either trained from scratch or fine-tuned a ResNet-50 \citep{he2016deep} model pre-trained on ImageNet-1k for 100 epochs using Adam with batch size 100. We linearly increased the learning rate for the first 5 epochs to $3 \cdot 10^{-3}$, in the case of training from scratch, or $3 \cdot 10^{-4}$, in the case of fine-tuning, and reduced it with the factor of 0.2 in the 50th, 70th, and 90th epochs. In both cases, we applied the following augmentations: resize to $292 \times 292$ pixels and random crop of a $256 \times 256$ pixels block, random flip left-right and up-down, and random rotation for $\left\{ 90, 180, 270, 360 \right\}$ degrees. At test time, the images were resized to $292 \times 292$ pixels and a $256 \times 256$ pixels block was cropped from the center. 

In the experiments with transfer learning, we used the pre-trained source models as either fixed feature extractors and trained a softmax classifier, or replaced the classification layer and fine-tuned the whole network on a target dataset. For target datasets, we used 20\% of images from each class for training and the remainder for testing. Following the usual protocol \citep{ericsson2021well}, we did not perform data augmentation when training the softmax and used the same augmentations as for the source datasets when fine-tuning the models. In the case of feature extraction data augmentation is omitted in order to assess the quality of the extracted features because they can also be used in different downstream tasks, such as image retrieval. It should be noted that a single dataset was not used as both the source and target dataset in the same experiment. 

For domain-adaptive pre-training, we fine-tuned ImageNet-1k pre-trained supervised and self-supervised ResNet-50 models on MLRSNet, as described previously, and used the resulting model as the pre-trained source model for transfer learning. In these experiments we did not use MLRSNet as a target dataset.

On the target datasets, we trained the softmax classifiers or fine-tuned the whole network for 100 epochs using Adam with batch size 100. For training the softmax we used a fixed learning rate $10^{-3}$ and for fine-tuning we used the same learning rate schedule as for fine-tuning the network pre-trained on ImageNet-1k. We report the classification accuracies on the test set for single-label tasks and F1-measures, with threshold 0.5, for multi-label tasks. All the networks were trained or fine-tuned on two Nvidia GTX 1080Ti GPUs with CUDA 11.0 and Intel Core i7-8700K CPU running Ubuntu 18.04.

We used the nonparametric bootstrap to estimate 95\% confidence intervals for each performance metric. We drew 1,000 replicates from the test set, and computed the performance metric on each replicate. This procedure produced a distribution for each metric, and we reported the 2.5 and 97.5 percentiles as a confidence interval.

\section{EXPERIMENTAL RESULTS}
\label{sec:experiments}

\subsection{Training from scratch vs. fine-tuning}

In the first experiment we compare training from scratch and fine-tuning the ImageNet-1k pre-trained network on remote sensing image datasets. In Table~\ref{tab:source-acc} the test accuracies/F1-measures of the models trained or fine-tuned on 80\% of the images from the used source datasets are reported. From these results we can observe that fine-tuning both supervised and self-supervised models pre-trained on ImageNet outperforms training from scratch in all the cases. However, for both variants of MLRSNet, as well as for PatternNet and RSI-CB, the differences are very small, and for RESISC45 the difference is around 2\%, indicating that, even for medium-sized datasets, ImageNet pre-training plays a diminishing role in HRRS scene classification. The differences between fine-tuned supervised and self-supervised models are not statistically significant. 

\begin{table*}[!htb]
\centering
\begin{tabular}{lccc}
\toprule
\multirow{2}{*}{Dataset} & \multicolumn{3}{c}{Training} \\ \cmidrule{2-4} 
                  & Scratch & Fine-tuning (supervised) & Fine-tuning (SwAV) \\ \midrule
   MLRSNet 
   (multi-label)  & 91.83 (91.69, 91.97) & 92.41 (92.27, 92.54) & \textbf{92.58} (92.45, 92.71)  \\ 
   MLRSNet
   (single-label) & 97.74 (97.55, 97.95) & 98.61 (98.46, 98.76) & \textbf{98.85} (98.70, 99.01) \\
   RESISC45       & 95.11 (94.56, 95.65) & \textbf{97.04} (96.57, 97.48) & 96.87 (96.44, 97.30) \\ 
   PatternNet     & 99.49 (99.31, 99.65) & \textbf{99.84} (99.74, 99.93) & 99.82 (99.70, 99.92) \\ 
   RSI-CB         & 99.39 (99.17, 99.60) & 99.55 (99.35, 99.72) & \textbf{99.64} (99.47, 99.80) \\ 
   AID            & 93.92 (92.85, 95.00) & 97.30 (96.55, 98.00) & \textbf{97.85} (97.20, 98.45) \\ 
   \bottomrule
\end{tabular}
\caption{Test accuracies/F1-measures (\%) with 95\% confidence intervals of models trained/fine-tuned on 80\% of the images from the HRRS datasets.}
\label{tab:source-acc}
\end{table*}

\subsection{Transfer learning}
\subsubsection{Feature extraction}
\label{sec:fex}

We now turn our attention to the transferability of representations learned on ImageNet-1k and HRRS image datasets. The results obtained by training a softmax classifier on the features extracted from each of the target datasets using the pre-trained models are shown in Table~\ref{tab:fex}. We can see that feature extractors pre-trained on HRRS image datasets outperform supervised feature extractors pre-trained on both ImageNet-100 and ImageNet-1k in all the cases. Interestingly, self-supervised pre-training considerably outperforms supervised pre-training on ImageNet-1k although in both cases domain-specific images are not used. Moreover, self-supervised pre-training is comparable to pre-training on remote sensing datasets, except in the case when RESISC45 is the target dataset.

When HRRS image datasets are used for pre-training, the best results are obtained using feature extractors pre-trained on MLRSNet, both single-label and multi-label, with multi-label pre-training winning in 3 out of 5 cases. In the tests on PatternNet and RSI-CB, single-label pre-training on MLRSNet marginally outperforms multi-label pre-training but the difference is very small and the performance is already saturated. Interestingly, pre-training on RESISC45 results in only slightly worse classification accuracies than pre-training on MLRSNet, despite being around three times smaller. Furthermore, pre-training on RESISC45 is comparable to supervised pre-training on ImageNet-1k and better than pre-training on ImageNet-100. Surprisingly, pre-training on PatternNet and RSI-CB yields much worse representations compared to all the other datasets. This was not expected having in mind that these datasets are comparable in sizes with RESISC45, significantly larger than AID, and with the classes similar to the classes in the target datasets. 

\begin{table*}[!htb]
 \centering
 \begin{tabular}{cLLccccc}
 \toprule
 \multirow{2}{*}{Source dataset} & \multicolumn{7}{c}{Target dataset} \\ \cmidrule{2-8} 
                   & MLRSNet (multi-label) & MLRSNet (single-label)       & RESISC45  & AID            & PatternNet     & RSI-CB         & UCM          \\ \midrule
   ImageNet-1k    & 83.77              & 91.69          & 86.94          & 90.81          & 98.63          & 98.50          & 92.86          \\
   ImageNet-100   & 81.23              & 88.22          & 82.10          & 87.21          & 98.16          & 97.81          & 90.12          \\
   ImageNet-1k (SwAV)    & \textbf{85.83}     & \textbf{93.22} & 89.21          & \textbf{92.98}          & \textbf{99.07}          & \underline{98.82}          & \underline{93.27}   \\
   MLRSNet
   (multi-label)   & -                  & -              & \textbf{93.21} & \underline{92.68} & 98.96          & 98.57          & \textbf{93.45} \\
   MLRSNet (single-label)        & -                  & -              & \underline{92.57}         & 91.09          & \underline{99.03} & \textbf{98.87} & 92.32          \\
   RESISC45  & \underline{85.19}     & \underline{91.96} & -              & 90.47          & 98.56          & 98.08          & 92.14          \\ 
   AID            & 80.83              & 85.78          & 79.81          & -              & 96.99          & 96.49          & 86.49          \\ 
   PatternNet     & 79.19              & 84.19          & 76.56          & 78.37          & -              & 97.10          & 83.57          \\ 
   RSI-CB         & 77.66              & 80.04          & 69.20          & 72.09          & 95.72          & -              & 74.82          \\ \bottomrule
 \end{tabular}
 \caption{Classification accuracies/F1-measures (\%) on 20\% of the images from the target datasets obtained by training softmax classifiers on the features extracted using the pre-trained networks. The best result for each target dataset is given in bold, and the second best is underlined.}
 \label{tab:fex}
\end{table*}

An intuitively plausible reason for good performance of the representations learned on HRRS datasets is that the source and target datasets contain the same or similar classes. To investigate this assumption we split MLRSNet into two subsets, one containing the classes present in UCM and the other with the rest of the classes. The sizes of these subsets are 50,197 and 58,964 images, respectively. We also made a third MLRSNet subset containing all the classes and half (54,573) the images. We train networks on all three subsets and use them to extract features from UCM. The resulting classification accuracies are given in Table~\ref{tab:ucm_classes}.

\begin{table}[!htb]
    \centering
    \begin{tabular}{cc}
        \toprule
        Subset & Accuracy (\%) \\ \midrule
        Same classes & 91.79 \\
        Different classes & 89.46 \\
        All classes & 92.26 \\ \bottomrule
    \end{tabular}
    \caption{Classification accuracies on UCM when different subsets of MLRSNet are used for training the feature extractor.}
    \label{tab:ucm_classes}
\end{table}

We observe that pre-training the feature extractor on the subset with the classes present in UCM results in better classification accuracy compared to pre-training on the subset with different classes. Moreover, pre-training on the subset with the same classes results in higher accuracy compared to training on, significantly larger, ImageNet-100. On the other hand, when the subset with different classes is used, the performance drop is less than 2\% compared to the subset with the same classes and around half percent compared to ImageNet-100. These results suggest that the performance indeed benefits from pre-training on the same or similar classes as in the target task.

When the subset with all the classes is used, the classification accuracy is additionally improved and is similar to the accuracies obtained by pre-training on RESISC45, full single-label MLRSNet, and supervised ImageNet-1k showing that in addition to domain similarity and overlap between the classes, class diversity also plays a role in training good feature extractors.

\subsubsection{Fine-tuning}

The classification accuracies obtained by fine-tuning the pre-trained networks on 20\% of the images from the target datasets are given in Table~\ref{tab:ft}. For comparison, we also report the classification accuracies obtained by training ResNet-50 on the target datasets from scratch. We see that both supervised and self-supervised pre-training on ImageNet-1k outperform training from scratch and pre-training on HRRS datasets in all the cases, with pre-training on multi-label MLRSNet being worse by around 1\%. Supervised and self-supervised pre-training on ImageNet-1k result in comparable performances, with self-supervised pre-training having a slight advantage. These results suggest that both the number of images and class diversity in ImageNet-1k contribute to obtaining a good initialization for fine-tuning in spite of the domain gap between ImageNet-1k and remote sensing images. Nevertheless, it should be noted that the differences are small challenging again the role of ImageNet-1k as an ubiquitous pre-training dataset. 

\begin{table*}[!htb]
 \centering
 \begin{tabular}{cLLccccc}
 \toprule
 \multirow{2}{*}{Source dataset} & \multicolumn{7}{c}{Target dataset} \\ \cmidrule{2-8} 
                  & MLRSNet (multi-label) & MLRSNet (single-label) & RESISC45 & AID   & PatternNet & RSI-CB & UCM   \\ \midrule
   Scratch        & 89.19                 & 93.87   & 85.44         & 79.14 & 98.04      & 97.29  & 58.93 \\  
   ImageNet-1k    & \underline{90.53}        & \underline{96.62}         & \underline{93.85}         & \underline{94.40} & \underline{99.51}      & \underline{99.15}  & \textbf{94.64} \\ 
   ImageNet-100   & 88.35                & 93.79   & 88.99         & 90.95 & 99.03      & 98.78  & 87.86 \\ 
   ImageNet-1k (SwAV) & \textbf{90.81}       & \textbf{97.27}   & \textbf{94.48}         & \textbf{95.37} & \textbf{99.65}      & \textbf{99.25}  & \underline{94.29} \\ 
   MLRSNet
   (multi-label)   & -                    & -       & 93.75         & 93.60 & 99.19      & 99.00  & 93.81 \\ 
   MLRSNet
   (single-label)  & -                    & -       & 92.17         & 92.16 & 99.11      & 98.90  & 92.50 \\ 
   RESISC45  & 89.08                & 94.53   & -             & 91.45 & 99.01      & 98.66  & 92.14 \\ 
   AID            & 88.46                & 93.60   & 88.52         & -     & 98.93      & 98.42  & 88.27 \\ 
   PatternNet     & 87.71                & 91.83   & 83.76         & 83.28 & -          & 98.13  & 86.73 \\ 
   RSI-CB         & 86.57                & 90.26   & 80.29         & 79.19 & 98.02      & -      & 78.39 \\ 
   \bottomrule
 \end{tabular}
 \caption{Classification accuracies/F1-measures (\%) on 20\% of the images from the target datasets obtained by fine-tuning the pre-trained networks. The best result for each target dataset is given in bold, and the second best is underlined.}
 \label{tab:ft}
\end{table*}

Pre-training on both variants of MLRSNet outperforms training from scratch in all the cases, and pre-training on RESISC45 is worse than training from scratch only in the case when the target task is multi-label MLRSNet. Furthermore, pre-training on both variants of MLRSNet, as well as on RESISC45 and AID, outperforms pre-training on ImageNet-100 on all the target datasets except PatternNet and RSI-CB, where the differences are very small.

Similarly to the feature extraction case, when PatternNet and RSI-CB are used for pre-training the obtained results are worse than in all the other cases. Moreover, for these two source datasets, the classification accuracies on most of the target datasets are not improved compared to training from scratch. A notable exception is UCM with too few training images for training a classifier from scratch. This result is in line with~\citep{pires2020convolutional}. 

By comparing the results in Table~\ref{tab:fex}. and \ref{tab:ft}. we see that fine-tuning the networks pre-trained on MLRSNet and RESISC45 for smaller target datasets only slightly improves classification performances in comparison with training only a classifier on the features extracted using the pre-trained network. On the other hand, in both experiments with MLRSNet as the target dataset, fine-tuning the network pre-trained on other HRRS source datasets results in improved classification performance compared to training only a softmax classifier. However, it should be noted that these fine-tuned networks exhibit similar performances as the networks trained from scratch on both MLRSNet tasks. Interestingly, fine-tuning the networks pre-trained on both ImageNet-100 and ImageNet-1k in all the cases considerably improves classification accuracies compared to the feature extraction case. 

To examine the impact of the class overlap between the source and target datasets on the fine-tuning performance we fine-tune the networks pre-trained on the subsets of MLRSNet on UCM. The obtained results are shown in Table~\ref{tab:ucm_classes_ft}. We can see that the obtained classification accuracies do not differ much compared to the feature extraction case. The improvements are at most half percent and the classification accuracies are comparable with those obtained by pre-training on RESISC45 and full single-label MLRSNet. Similarly to the feature extraction case, pre-training on the subset with the same classes outperforms pre-training on the subset with different classes. However, more diversity in the subset with all MLRSNet classes in this case does not bring additional improvement.

\begin{table}[!htb]
    \centering
    \begin{tabular}{cc}
        \toprule
        Subset & Accuracy (\%) \\ \midrule
        Same classes & 92.38 \\
        Different classes & 89.88  \\
        All classes & 92.38  \\ 
        \bottomrule
    \end{tabular}
    \caption{Fine-tuning accuracies on UCM when different subsets of MLRSNet are used for pre-training.}
    \label{tab:ucm_classes_ft}
\end{table}

\subsection{Domain-adaptive pre-training}

The results in Section~\ref{sec:fex} show that, when pre-trained models are used as feature extractors, in-domain pre-training is comparable to self-supervised ImageNet pre-training and both approaches outperform supervised ImageNet pre-training. Furthermore, all the approaches result in similarly performing fine-tuned models. Therefore, in order to leverage the advantages of both ImageNet and in-domain pre-training we investigated the quality of the representations obtained by domain adaptive (DA) pre-training, i.e. fine-tuning the network pre-trained on ImageNet-1k using an in-domain dataset different from the target dataset and using the resulting model as a feature extractor or additionally fine-tuning it for the target task. In this section, we used MLRSNet (single and multi-label variants) as the source dataset and RESISC45, AID, and UCM as the target datasets. We excluded PatternNet and RSI-CB from these experiments because their performances had already been saturated in the previous experiments and pre-training on these datasets did not bring improvements on the target tasks compared to training from scratch. In all the cases we used supervised fine-tuning on MLRSNet. To get better insight into the impact of class overlap between the dataset used for domain adaptation and the target dataset we also performed DA pre-training using the subsets of MLRSNet containing the same and different classes as UCM and evaluated the domain-adapted models on UCM classification. 
 
\subsubsection{Feature extraction}

The results obtained by training softmax classifiers on the features extracted using domain-adapted models are given in the upper half of Table~\ref{tab:da} (marked with FE). In comparison with the results in Table~\ref{tab:fex} we can see that DA pre-training improves the classification accuracies on the target datasets compared to using only ImageNet-1k or MLRSNet pre-training. The improvement is present when both supervised and self-supervised ImageNet pre-training are used, with self-supervised pre-training outperforming supervised. In contrast to pre-training only on MLRSNet, in this case the multi-label variant of MLRSNet does not show any advantages over the single-label variant. 

\begin{table*}[!htb]
 \centering
 \begin{tabular}{ccccc}
 \toprule
 \multirow{2}{*}{Pre-training} & \multirow{2}{*}{Transfer} & \multicolumn{3}{c}{Target dataset} \\ \cmidrule{3-5} 
                                 &                           & RESISC45 & AID   & UCM   \\ \midrule
   ImageNet (supervised) $\rightarrow$ MLRSNet (multi-label) & FE                        & 94.50             & 92.51          & 94.35 \\
   ImageNet (supervised) $\rightarrow$ MLRSNet (single-label) & FE                        & 94.69 & 92.99 & 94.17 \\
   ImageNet (SwAV) $\rightarrow$ MLRSNet (multi-label) & FE & 94.54 & 93.29 & 94.96 \\
   ImageNet (SwAV) $\rightarrow$ MLRSNet (single-label) & FE & \textbf{95.24} & \textbf{93.92} & \textbf{96.89}  \\
   \midrule
   ImageNet (supervised) $\rightarrow$ MLRSNet (multi-label) & FT                        & 94.27             & 94.31          & 93.87 \\ 
   ImageNet (supervised) $\rightarrow$ MLRSNet (single-label) & FT                        & 95.14    & 95.54 & 95.12 \\ 
   ImageNet (SwAV) $\rightarrow$ MLRSNet (multi-label) & FT & 95.49 & \textbf{96.17} & 96.55 \\
   ImageNet (SwAV) $\rightarrow$ MLRSNet (single-label) & FT & \textbf{95.89} & 96.09 & \textbf{97.14} \\
   \bottomrule
 \end{tabular}
 \caption{Classification accuracies (\%) on the target datasets obtained by transfer learning using the domain-adapted models. FE denotes using domain-adapted models for feature extraction and FT fine-tuning the domain-adapted model.}
 \label{tab:da}
\end{table*}

The classification accuracies obtained by using the feature extractors domain-adapted on different subsets of MLRSNet are given in Table~\ref{tab:da_ucm_classes}. We can see that DA pre-training on the MLRSNet subset with the same classes as in UCM outperforms DA pre-training on the subset with the different classes by around 4\% and achieves almost the same results as when the MLRSNet subset with all the classes and half the images is used. However, when the MLRSNet subset with different classes is used for DA pre-training, the classification accuracies on UCM are lower compared to ImageNet-1k pre-training only. Apparently, DA pre-training shifted the features towards the discrimination between the classes not present in UCM reducing their performance on UCM classification. These results suggest that feature extractors can benefit from DA pre-training in those cases when there exists class overlap between the dataset used for DA pre-training and the target dataset. 

\begin{table}[!htb]
    \centering
    \begin{tabular}{cc}
        \toprule
        Subset & Accuracy (\%) \\ \midrule
        Same classes & 95.18 \\
        Different classes & 91.25 \\
        All classes & 95.30 \\ \bottomrule
    \end{tabular}
    \caption{Classification accuracies on UCM when different subsets of MLRSNet are used for DA training of the feature extractor.}
    \label{tab:da_ucm_classes}
\end{table}

\subsubsection{Fine-tuning}

The results obtained by fine-tuning the domain-adapted models on the target datasets are given in the lower half of Table~\ref{tab:da} (marked with FT). In comparison with Table~\ref{tab:ft}, we can see that DA pre-training improves the classification accuracies compared to both supervised and self-supervised pre-training on ImageNet-1k only. Moreover, the self-supervised model makes for a better basis for DA pre-training than the supervised one.

Fine-tuning the models obtained by DA pre-train\-ing on the different subsets of MLRSNet results in the classification accuracies given in Table~\ref{tab:da_ucm_classes_ft}. From these results similar conclusions as in the case of feature extraction can be drawn. Specifically, DA pre-training on the MLRSNet subset with the same classes as in UCM is considerably better than pre-training on the MLRSNet subset with the different classes. In this case, it is even better than DA pre-training with all the classes and half of the training images. Furthermore, it outperforms pre-training on both ImageNet-1k and MLRSNet (multi-label). Therefore, when the models are fine-tuned for the target task, DA pre-training using a dataset with class overlap with the target dataset is beneficial. 

\begin{table}[!htb]
    \centering
    \begin{tabular}{cc}
        \toprule
        Subset & Accuracy (\%) \\ \midrule
        Same classes & 95.77 \\
        Different classes & 92.44  \\
        All classes & 94.70 \\ 
        \bottomrule
    \end{tabular}
    \caption{Fine-tuning accuracies on UCM when different subsets of MLRSNet are used for DA pre-training.}
    \label{tab:da_ucm_classes_ft}
\end{table}

\section{DISCUSSION}
\label{sec:discussion}

In summary, our experimental results show that training from scratch on most of the used HRRS image datasets results in only slightly lower performance than fine-tuning the ImageNet pre-trained models on the same datasets. For example, the differences on both variants of MLRSNet, PatternNet and RSI-CB are less than 1\%, with somewhat larger gaps on RESISC45 and AID. These results suggest that for larger and some medium-sized HRRS image datasets we might avoid ImageNet pre-training and still achieve competitive results. 

When pre-trained networks are used as fixed feature extractors, pre-training on HRRS image datasets outperforms supervised pre-training on ImageNet. Clearly, the features obtained using the networks pre-trained on HRRS images are better suited for other HRRS scene classification tasks than the features computed using the ImageNet pre-trained network. However, when a self-supervised model is used as a feature extractor, the performances are similar to pre-training on HRRS image datasets. On the other hand, when pre-trained networks are end-to-end fine-tuned, supervised ImageNet pre-training slightly outperforms pre-training on HRRS image datasets, while self-supervised ImageNet models after fine-tuning outperform both supervised ImageNet and in-domain models. 

When PatternNet and RSI-CB are used as source datasets in both transfer learning scenarios, the obtained classification accuracies are lower compared to pre-training on other datasets. However, the classification accuracies on PatternNet and RSI-CB exceed 99\%, showing that the performance on a source task is not always a good predictor of the performance on a target task. It is possible that classification of PatternNet and RSI-CB is too easy, which prevents the network from learning useful features for other HRRS scene classification tasks. 

The recent experiments on Million-AID \citep{long2022aerial} showed that a large HRRS image dataset can outperform ImageNet as a source dataset in a transfer learning scenario and our results indicate that the gap between the models trained from scratch and fine-tuned models pre-trained on ImageNet is very narrow even for medium-sized HRRS image datasets. Nevertheless, there is another benefit of ImageNet pre-training. The models pre-trained on ImageNet and fine-tuned on a HRRS image dataset, i.e. domain-adaptive pre-trained models, outperform the models pre-trained only on either ImageNet or MLRSNet in both transfer learning scenarios. Nevertheless, although the improvements can be readily achieved by domain-adaptive pre-training on medium-sized datasets, our results suggest that it is important that there exists class overlap between the dataset used for domain adaptation and the target dataset.

In Table~\ref{tab:sota} the results on RESISC45 and AID obtained using DA pre-training are compared to recent state-of-the-art HRRS scene classification methods. Two of the methods, ResNet50+EAN \citep{zhao2020remote} and PCNet\citep{zhang2021pairwise} use the same ResNet-50 backbone as in our experiments, while GLDBS \citep{xu2021remote} uses ResNet-34. The best results obtained using pre-training on Million-AID used DenseNet-169 and ResNet-101 for classification of RESISC45 and AID, respectively. We can see that domain-adaptive pre-training without any additional modifications of the model is able to surpass the classification accuracies achieved using the recently proposed methods, as well as pre-training on an order of magnitude larger Million-AID. Therefore, DA pre-training can serve as a strong baseline for HRRS scene classification. Furthermore, DA pre-training is orthogonal to other extensions of the ResNet architecture proposed in the literature and we believe that it is possible that their combination would lead to further increase in classification accuracy.

\begin{table}[!htb]
 \centering
 \begin{tabular}{ccc}
 \toprule
 \multirow{2}{*}{Method} & \multicolumn{2}{c}{Target dataset} \\ \cmidrule{2-3} 
                                                  & RESISC45          & AID  \\ \midrule
   ResNet50+EAN \citep{zhao2020remote}             & 93.51             & 93.64 \\
   GLDBS \citep{xu2021remote}                      & 94.46             & 95.45 \\
   PCNet \citep{zhang2021pairwise}                 & 94.59             & 95.53 \\
   Million-AID \citep{long2022aerial}              & 94.26             & 95.40 \\ 
   Domain-adaptive pre-training (ours)            & \textbf{95.89}    & \textbf{96.17} \\
   \bottomrule
 \end{tabular}
 \caption{Comparison with the state-of-the-art methods.}
 \label{tab:sota}
\end{table}

\section{CONCLUSION}
\label{sec:conclusion}

In this paper we empirically showed that, although ImageNet pre-training in its traditional form gradually loses its appeal, we can obtain additional improvements by using a second round of pre-training using in-domain data i.e., domain-adaptive pre-training. Therefore, the answer to the question posed in the title is positive and we can still benefit from ImageNet pre-training by coupling it with domain-adaptive pre-training. Additionally, since self-supervised pre-training is on par or better than supervised pre-training, for pre-training we only need the images from ImageNet and not their labels. An important consequence of our work is that domain-pretrained networks can be used as backbones in all applications where supervised ImageNet pre-trained networks have traditionally been used.

In the future work we plan to further investigate what makes a dataset suitable for pre-training, as well as why fine-tuning the network pre-trained on ImageNet outperforms fine-tuning the networks pre-trained on HRRS image datasets in spite of the better results of the latter as feature extractors. Another interesting question is the impact of the choice of the network layers for fine-tuning on the classifier performance. Finally, transferability of representations to other target tasks, such as object detection and semantic segmentation is also an interesting direction of research.

\section*{ACKNOWLEDGEMENTS}\label{ACKNOWLEDGEMENTS}

We thank Mihajlo Savi\'{c} for insightful discussion. Valuable comments from the reviewers are gratefully acknowledged.

This research was funded in part by the European Commission under the Horizon 2020 European research infrastructures grant agreement no. 857645 (NI4OS-Europe project) and in part by the Ministry of Scientific and Technological Development, Higher Education and Information Society of the Republic of Srpska under contract 19.032/961-102/19, ``Service for Classification of Remote Sensing Images''.

{
	\begin{spacing}{1.17}
		\normalsize

	\end{spacing}
}

\end{document}